\documentclass{article}

 \usepackage[preprint]{neurips_2025}

\usepackage[utf8]{inputenc} 
\usepackage[T1]{fontenc}    
\usepackage{hyperref}       
\usepackage{url}            
\usepackage{booktabs}       
\usepackage{amsfonts}       
\usepackage{nicefrac}       
\usepackage{microtype}      
\usepackage{xcolor}         
\usepackage{graphicx}
\definecolor{airforceblue}{rgb}{0.36, 0.54, 0.66}
\definecolor{bluegray}{rgb}{0.4, 0.6, 0.8}
\definecolor{bleudefrance}{rgb}{0.19, 0.55, 0.91}
\hypersetup{colorlinks,linkcolor={airforceblue},citecolor={bleudefrance},urlcolor={bluegray}}

\makeatletter
\newcommand*{\rom}[1]{\expandafter\@slowromancap\romannumeral #1@}
\makeatother

\title{Ten Principles of AI Agent Economics}

\author{%
  Ke Yang, ChengXiang Zhai \\
  University of Illinois Urbana-Champaign\\
  \texttt{\{key4, czhai\}@illinois.edu} \\
}

\begin{document}

\maketitle

\begin{abstract}
The rapid evolution of AI-based autonomous agents is reshaping human society and economic systems, as these entities increasingly demonstrate human-like or even superhuman intelligence. From mastering complex games like Go to solving diverse general-purpose tasks with large language and multimodal models, AI agents are transitioning from narrowly focused tools to versatile participants in economic and social ecosystems. Their autonomy and decision-making capabilities are poised to impact industries, professions, and human lives profoundly, raising critical questions about their integration into economic activities, potential ethical concerns, and the balance between their utility and safety.  

As a step toward answering those questions, this paper outlines ten principles of AI agent economics, providing a foundational framework for understanding how AI agents make decisions, influence social interactions, and participate in the broader economy. Drawing on insights from economics, decision theory, and ethics, we explore fundamental questions, such as whether AI agents could evolve beyond tools into independent entities, how their participation reshapes labor markets, and what ethical constraints are necessary to ensure their alignment with human values. The principles we propose complement existing economic theories while addressing the unique characteristics of AI agents, offering a roadmap for their responsible integration into human systems.  

Beyond theoretical contributions, this paper emphasizes the importance of future research in trustworthiness in AI applications, robust ethical guidelines and regulatory oversight about the deployment of AI agents. As we stand at the cusp of a transformative era, this work serves as both a guide and a call to action for ensuring that AI agents contribute positively to human progress while mitigating potential risks associated with their unprecedented capabilities.
\end{abstract}
\section{Introduction}

The discourse on AI-based autonomous agents reshaping our world intensifies with each passing day. These AI agents are computer programs exhibiting human-like or even superhuman intelligence, granted a degree of autonomy to perform tasks that previously required human intervention, either semi-autonomously or fully autonomously. The ambition to expand the reach of AI agents is fueled by the success of high-performance programs: In the 2010s, AlphaGo defeated human expert Go players, with the pinnacle of expertise in Go being dominated by algorithms \citep{silver2016mastering}; in the 2020s, large language models emerged, demonstrating that when computational scales are significantly increased, programs can solve general-purpose problems—such as answering questions, writing code, and solving math problems—akin to humans \citep{brown2020languagemodelsfewshotlearners}. Language, a vast and complex space of action once used by humans to claim superior intelligence in nature, has been mastered by these models. Within less than a decade, foundational models for vision, audio, and other multimodal tasks have emerged, and despite their limitations, they have matched or even surpassed human capabilities in fundamental understanding and generation tasks \citep{rombach2022highresolutionimagesynthesislatent,borsos2023audiolmlanguagemodelingapproach,radford2021learningtransferablevisualmodels}.

It is foreseeable that building on these foundational models with domain adaptation training for specific tasks, 
a wide spectrum of AI agents will be developed quickly in all kinds of application areas. 
Indeed, many industries already employ AI agents, and more will follow: we foresee agents providing tailored guidance for agriculture or livestock based on geographic conditions, controlling or designing algorithms to regulate furnace temperatures with minimal human intervention, appearing as physical entities in classrooms, clinics, logistics, and retail, and contributing to technology, creativity, and environmental protection.

Supporters of AI are eager to see these meticulously designed and trained programs replace many human roles in social divisions of labor—some envision AI leading the fourth industrial revolution, where highly repetitive and tedious tasks are handled by agents, optimizing social functions, saving human labor, and thus yielding greater returns for the human community at lower costs. However, numerous works of fiction portray the future of AI development as catastrophic, with popular discussions on scenarios like \textit{The Matrix}, where humans are used as programs' energy sources \citep{the_matrix}, or \textit{I, Robot}, where robots' ``human-protective plans'' lead to restrictions on human freedom \citep{irobot}. If these imagined evolutions of the future are plausible regarding the conditions for the evolution of history, they indicate one possibility of coexistence between humans and agents. In any case, public discourse inevitably points to this: Given the impressive intelligence and potential productivity gains of AI agents, they will undoubtedly be granted autonomy, integrated into human production and life, and participate in economic activities.

As they are in their nascent stage, we must figure out how to balance the autonomy and safety of AI agents, ensuring that at least the fundamental first steps are taken correctly. Many questions that once appeared to hold little relevance to our society have now become both significant and urgently pressing: Do AI agents have consciousness or needs? Will they always remain humans' tools, or could they become citizens? Could there be AI agents designed specifically to cause harm? Does a system with several unmanned reconnaissance drone duplicates but a central decision-making unit constitute one agent or multiple agents? Can workers and graduate students delegate their responsibilities to AI agents and take leave? What about judges and government officials? Can human occupations be entirely replaced by AI agents? Will AI agents destroy human civilization?
  
\textbf{Answers to such questions would only be possible through a scientific analysis of the decision-making mechanisms of AI agents and an assessment of their behaviors in economic and social contexts, both as individual agents and as members of an economic ecosystem that integrates AI systems and human participants. To this end, we offer a perspective from participants in AI systems development in this paper, outlining principles and fundamental facts regarding AI agents' involvement in human economic activities. This includes how AI agents make decisions, how they affect other intelligent entities in social interactions, and how the economy functions with AI agent participation.} The AI agent economics principles we put forward offer objective insights about agent decision-making. 
Based on these principles, we can make reasonable estimates of the agents' decision-making process and impacts as they engage in economic activities in the future, thus facilitating research prioritization and policy making in the future.

It is important to note that the AI agent economics principles we propose are intended to complement, rather than replace, existing human economics principles, which remain applicable in the broader context of human and AI agent interactions. General principles, such as facing trade-offs and responding to incentives \citep{mankiw2021principles}, apply equally to the behavior of AI agents. Moreover, research into AI agent decisions will benefit from existing studies on human decision-making, such as asymmetric information and game theory \citep{auronen2003asymmetric,fudenberg1991game}.

Finally, \textbf{we want to emphasize AI agent ethics}, which we will also cover in our principles. Pessimistic scientists caution against broadcasting Earth's sounds into space due to finite resources and potential inter-civilizational competition or upheaval. In parallel, when building AI agents, we are creating another intelligence that rivals or surpasses human intelligence, potentially cultivating another form of civilization. \textbf{As the old adage goes, \textit{A single thought can make a Buddha or a demon}, at this transformative starting point, the human community must reach a consensus on the absolute ethical principles AI agents must adhere to, constraining academic and industrial agent development and urging legislative and governmental regulation.}

The rest of the  paper is organized into three sections: \textit{i)} How AI agents make decisions, \textit{ii)} How agents influence other intelligent entities participating in social activities, and \textit{iii)} How the overall economy functions with agent participation. Each section proposes corresponding AI agent economics principles and provides explanations and arguments. After the main discussion, we will revisit the earlier questions guided by our principles. \textbf{Through this paper, we aim to equip a broader audience with the essential background knowledge for understanding the development of AI agents and their future involvement in human life, aiding their decision-making, such as purchasing stocks or home robots. For ML researchers, we summarize the objective patterns and risks in the development of AI agents, providing a roadmap to help them track progress and avoid pitfalls in algorithm design. For everyone else, beyond providing information, we emphasize that trustworthiness is as crucial as performance, efficiency, and cost in AI applications, concerning human development in the presence of another similar or superior intelligence.}
\section{How AI Agents Make Decisions}
\subsection{Principle \rom{1}: The fundamental structure of AI agents differs from that of humans, leading to distinct decision-making drivers and mechanisms.}

\paragraph{For decision-making drivers, AI agents learn and decide by optimizing designated objective functions.}
This creates a unique contrast: AI cognition relies on closed-form mathematical models established before deployment (\textit{theory into practice}), whereas human cognition must be reverse-engineered from observed brain function (\textit{practice into theory}). With access to model parameters, AI decision behaviors become largely predictable and reproducible. Although often deemed “black-boxes,” this primarily reflects challenges in linking input-output behavior to the relevant parameters—paralleling the difficulty of mapping neural functions in human brains. Both computer science and biology actively research interpretability of these complex systems, yet AI’s input-output causality remains deterministic at its core. Furthermore, while humans pursue \textit{self-need-oriented} objectives, most current AI models are built for \textit{human-need-aligned pattern recognition}; in time, these \textit{pattern-recognition} goals may evolve into \textit{need-driven} ones, prompted by rising demands for autonomous, self-evolving AGI agents.

\paragraph{For decision-making mechanism, AI agents diverge from humans in many physiological respects: they bypass hormone-driven emotions, biological vulnerabilities, sleep or nutrition needs, are immune to aging or disease, and are unbounded by  built-in memory limits and the body-brain coupling, to name a few.}
In early AI research, a \textit{human-likeness as virtue} paradigm guided development, using human cognition as both benchmark and blueprint. This anthropomorphizing overlooks key differences, for instance, AI lacks the hormonal basis for creativity and passion, and its “neural” structure can be replaced by updated mathematical models rather than awaited through generational shifts. Such structural distinctions critically affect how self-needs form and how learning trajectories are collected. As AI grows more autonomous, it may no longer be confined to human-aligned needs, implying that humans and AI could eventually develop into separate populations with divergent interests and behaviors. Further study is essential to anticipate and understand the traits and dynamics of this emergent intelligent population.

\begin{figure*}[ht]
    \centering
    \includegraphics[width=\textwidth]{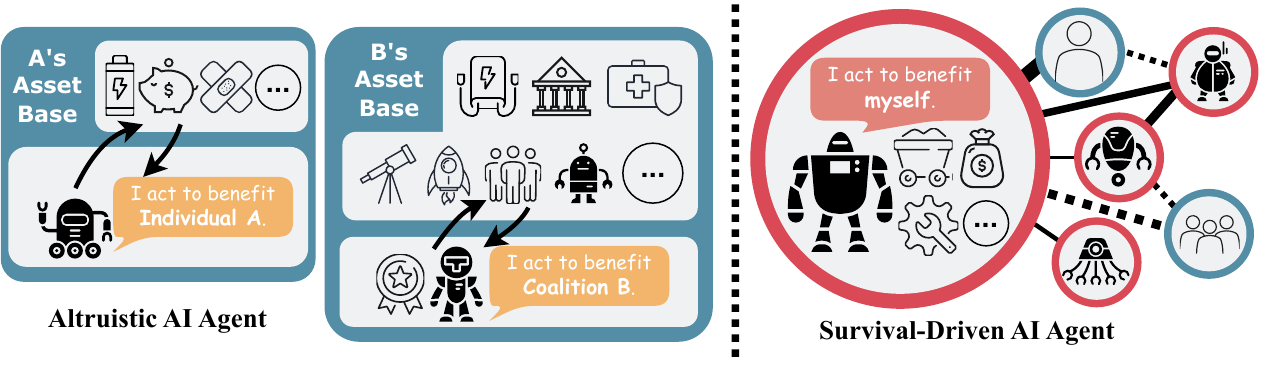}
    \vspace{-2em}
    \caption{\label{fig:ai_decide}
        Two prototypes of AI agents' goal designs: to help humans, or to survive.
    }
    \vspace{-1em}

\end{figure*}

\subsection{Principle \rom{2}: The decision-making processes of AI agents are grounded in the formation of their self-awareness and self-needs.}

\paragraph{The formation of an AI agent’s self-awareness necessitates continuous environmental perception, feedback, and memory retention.}
This principle extends to widely recognized chat assistants, which the public and academia often question regarding their self-awareness—encompassing elements such as existence, consciousness, identity, and emotions. Although these assistants exhibit sophisticated capabilities, they lack a crucial human-like facet of intelligence: true self-awareness. This shortcoming stems largely from their training and inference paradigms, where models remain dormant until prompted by a user, then terminate upon session conclusion. Such interactions primarily extract information for human needs, rather than fostering the agent’s identity formation. Further, unless curated as future training data, these interactions carry no lasting impact on the model’s parametric understanding. Consequently, \textit{dormant cognition}, \textit{instrumental intelligence}, and \textit{memory fragmentation} prevent ongoing perception, robust identity feedback, and continuity in memory—limitations that could be overcome by modified training paradigms enabling more persistent experiences and continuous self-reflection.

\paragraph{An AI agent’s self-needs are expressed in the objective functions it optimize—mathematical descriptions of its goals.}
Current mainstream \textit{human-need-aligned} objective functions effectively translate to “AI optimizes to assist humans.” For example, large language models that employ causal language modeling can be seen as ``reciting a human corpus,'' driven by the human requirement for extensive knowledge memorization and generalization. From the AI’s perspective, fulfilling this core objective function equates to satisfying its self-needs.

\paragraph{An AI agent’s learning and decision-making mode will evolve from pattern-recognition-driven to goal-driven paradigms.}
Today’s \textit{pattern-recognition-driven} systems focus on tasks ranging from text and image generation to multi-modal data alignment, guided by human-defined model architectures, data strategies, and objective functions. Future general-purpose AI agents, however, will require minimal human intervention, transitioning toward endogenously \textit{goal-driven} systems. Under this approach, humans would specify only broad objectives—such as safe human progress or societal welfare—while AI agents would devise sub-objectives, determine self-evaluation metrics, pursue resources and feedback from their environment, and selectively update their parameters to improve. This transformation signifies a pivotal step toward genuinely autonomous learning.

\paragraph{Self-awareness provides conditions for AI-agent-specific decision-making, while self-needs determine goals, learning processes, and outcomes.}
Consider a public policy scenario requiring assessment of whether to clear a green space for constructing a large mobile charging station. Self-aware AI agents would approach this decision differently from humans, as they need not consider factors like air quality or residential comfort that concern biological entities. Instead, such agents would recognize that expanded charging infrastructure significantly benefits their operational flexibility in the region. Furthermore, the programmed self-needs of these agents determine their decision orientation: AI agents programmed to prioritize human welfare would potentially evaluate the policy based on sustainability impacts and long-term human benefits, whereas those optimized for short-term economic gains would likely prioritize the new facility's revenue potential.

\subsection{Principle \rom{3}: Most AI agents will exist as human proxies, with their guiding principles intrinsically tied to the interests of the individuals or coalitions they represent.}

\paragraph{Altruistic AI agents will make decisions to satisfy human needs and advance human welfare.}
As shown in the left panel of Figure \ref{fig:ai_decide}, these agents function as assets within an owner’s resource base and operate as intelligent tools with strictly delegated access. Their decision outputs benefit their owners rather than themselves, and their entire operational cycle—activation, execution, energy supply, maintenance, and eventual decommissioning—remains under owner control. Many current AI systems, including household robots and enterprise AI implementations, follow this principle. It is likely that such \textit{instrumental intelligence} will remain the dominant AI paradigm across economic domains. Depending on the degree of self-awareness these agents display and on public empathy, future debates may arise regarding their social standing and potential civil and property rights.

\paragraph{AI agents emerging from malign groups will defend dark faction interests.}
Because AI agents’ underlying configurations are malleable, illicit operations—ranging from trafficking to fraud—can develop \textit{proficient malevolent actors} akin to legitimate corporate AI workers. From both ethical and economic perspectives, preserving their hardware for possible goal-system recalibration represents a viable rehabilitation approach, rather than outright destruction of these compromised AI systems.

\paragraph{AI agents with survival-driven goals could embody the dystopian futures depicted in allegory.}
As depicted in the right panel of Figure \ref{fig:ai_decide}, such survival-driven machines would prioritize self-sustaining capabilities, including energy acquisition and system optimization. In contrast to altruistic AI, they would exist as independent entities equipped with their own resources, potentially forming separate organizational structures or societies and competing with humans. Given their limited economic utility and considerable risks, these survival-centric designs are unlikely to predominate. Accordingly, the subsequent analysis in this paper assumes that most AI agents will be aligned with human goals.

\subsection{Principle \rom{4}: AI agent decision-making can be framed as a constrained optimization problem, where autonomy is one of the key parameters affecting operational efficiency.}

\paragraph{Autonomy quantifies the scope and limits of an AI agent’s decision-making authority, representing a primary constraint beyond standard resource limitations.}
It is important to distinguish between autonomy as a capability and autonomy as a right. The former denotes the agent’s underlying abilities—such as its knowledge base and acquired experience—enabling it to make sound decisions with minimal oversight. The latter refers to whether the AI agent is granted the authority to allocate resources and implement decisions independently. We focus on autonomy as a right, since limiting it restricts the agent’s freedom to self-direct. Moreover, assigning decision-making responsibility to an AI agent entails that human operators implicitly accept accountability for its actions. For example, when a vehicle owner fully entrusts driving to an AI driver, they agree to bear the consequences of any errors. Likewise, permitting AI-based teachers, journalists, physicians, judges, or military systems to act without interruption signifies acceptance of responsibility for outcomes stemming from algorithmic shortcomings, hallucinations, biases, or susceptibility to adversarial attacks.

\paragraph{Constraints restrict the solution space, so AI agents with limited autonomy may produce suboptimal decisions; thus, a balance between decision quality and safety is crucial.}
For instance, AI assistants with broader access to a user’s personal information can deliver more tailored support but at the cost of heightened privacy concerns. Granting far-reaching permissions—such as access to schedules, browsing history, or location data—can enhance personalization while also amplifying ethical and security risks. Similarly, rigid adherence to Asimov’s First Law of Robotics (``a robot may not injure a human being'') would preclude AI-driven law enforcement from neutralizing dangerous threats, whereas granting weaponized systems full autonomy delegates targeting decisions entirely to algorithms, posing critical questions about reliability and ethical supervision.
\section{How AI Agents Influence Other Intelligent Participants in Social Activities}
\begin{figure*}[ht]
    \centering
    \includegraphics[width=\textwidth]{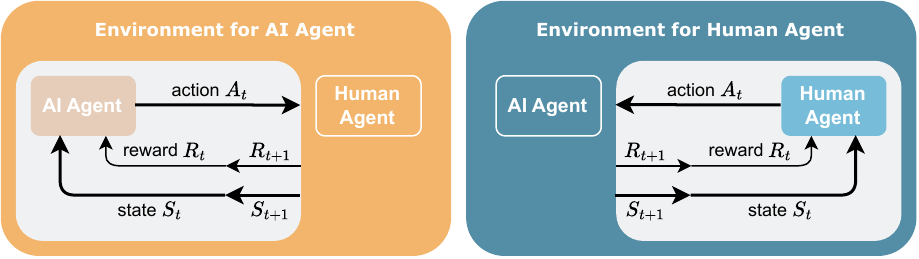}
    \vspace{-2em}
    \caption{\label{fig:ai_influence}
        The agent-environment interaction from \textit{AI/human-agent-centric} perspective.
    }
    \vspace{-1em}

\end{figure*}

\subsection{Principle \rom{5}: AI agents and human agents coexist in the same physical world, mutually influencing one another as interactive participants in shared environments.}

\paragraph{AI agents are transitioning from recipients of human knowledge to providers of novel learning materials for human benefit.}
Recent research increasingly frames AI design within interactive paradigms, such as reinforcement learning, in which agents adapt to feedback from their surroundings. Initially, this approach was AI-centric (Figure \ref{fig:ai_influence}, left panel), with humans supplying learning signals to guide AI behavior. This is because, early AI systems fell short of human-level intelligence, necessitating human-driven training loops focused on improving AI capabilities via behavioral constraints informed by human expertise. However, as AI systems advance toward superintelligence, we anticipate a shift (Figure \ref{fig:ai_influence}, right panel), whereby these agents will transcend current knowledge boundaries and produce insights that benefit human learning. In the short term, AI agents learning from humans remains beneficial; over time, humans learning from increasingly capable AI agents may prove equally valuable.

\paragraph{AI agents' decisions will systematically transform the material and intellectual dimensions of human existence.}
Materially, the impact will permeate all levels of society—exemplified by, but not limited to, AI agents actively participating in producing crops and essential consumer goods. Similarly, AI algorithms embedded in smart homes and intelligent urban planning represent how these technologies will influence individual daily experiences while simultaneously affect systemic sustainability metrics such as carbon emissions, water resource management, green space allocation, and overall urban habitability. Intellectually, AI agents can potentially influence human thoughts systematically through education and media, extending further than the already-documented concern of algorithmic filter bubbles. For examples, the proliferation of AI for education and potential widespread adoption of AI tutors means impressionable learners will have their knowledge foundations and value systems shaped by AI-engaged educational frameworks. Likewise, AI-generated media content may shape public discourse and cultural development.

\subsection{Principle \rom{6}: Cooperation and competition exist between AI agents, and between AI agents and humans, with decision-making aimed at serving the interests of their respective stakeholders.}

\paragraph{Under the assumption of altruistic AI agents, the coordination or conflict of decision-making goals reflects the alignment or divergence of human interests.}
In everyday settings, users might deploy instrumental AI to refine resumes, compete for limited educational or employment opportunities, or obtain high-demand event tickets. While such AI agents confer advantages on their users, they may simultaneously disadvantage rivals competing for the same resources. Evaluations of these interactions depend on whether the interests of each party align or conflict. This principle extends to broader contexts, such as corporate bidding or international negotiations. Ultimately, the morality of a tool lies not in the AI itself but in whether the human ambitions it serves are in harmony or at odds with those of others.

\paragraph{Under the assumption of survival-driven AI agents, humans and AI can be seen as two distinct species striving to coexist.}
Competing for finite resources, both groups act according to their own survival imperatives, with interactions governed by the principle of survival of the fittest. Whether contesting economic opportunities, decision-making authority, or computational resources, humans and AI agents seek to maintain relevance in environments requiring continual adaptation. Such competition may manifest as collaboration, conflict, or mutual adjustment, but it is ultimately driven by the need to secure the means for continued existence and growth.
\section{How the Overall Economy Operates with AI Agent Participation}

\begin{figure*}[ht]
    \centering
    \includegraphics[width=\textwidth]{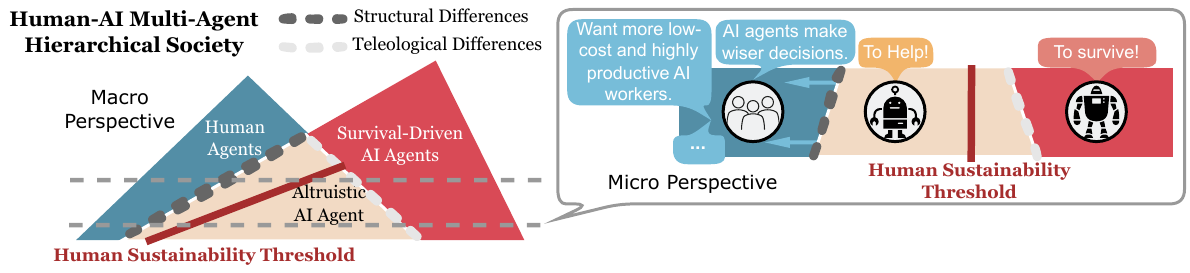}
    \vspace{-2em}
    \caption{\label{fig:ai_participate}
        The micro and macro perspective of how AI agents would get involved in the human society, economy, and power system. They will take increasing share for their ever-improving labor productivity and decision-making capability. 
    }
    \vspace{-1em}

\end{figure*}

\subsection{Principle \rom{7}: AI agents will exhibit functional specialization and hierarchical organization, seamlessly integrating into human social, economic, and power structures.}

\paragraph{AI agents' operational modes range from decentralized systems, each tasked with specific functions, to a singular centralized system that holistically optimizes conditions and makes global decisions.}
For instance, in intelligent transportation systems, localized in-vehicle AI agents can optimize individual trips, or a city-level system can coordinate overall traffic efficiency. The choice between these extremes depends on whether computation-intensive global optimization is preferable to targeted task delegation. Centralized systems demand significant modeling and computational resources, raising questions of efficiency, scalability, and necessity—e.g., should a superintelligent agent tasked with advancing human knowledge also regulate every home thermostat? Moreover, entrusting critical decisions to one AI can introduce systemic vulnerabilities if errors arise.

\paragraph{AI agents must function both as decentralized actors and comply with directives from higher-level, neutral supervision.}
Their differing comparative advantages—resource-intensive yet versatile general agents versus lightweight, specialized agents—naturally foster division of labor and stratification akin to human social structures, as shown in Figure \ref{fig:ai_participate}. Larger agents excel at strategic planning and coordination, while smaller ones efficiently execute specialized tasks. As AI agents integrate into human systems, their roles will adapt across multiple layers, largely conforming to established human rules. Thus, AI agents may act as executors, following high-level directives, or as decision-makers, independently reasoning, planning, and delegating tasks to sub-agents. This layered, collaborative approach foreshadows the emergence of roles such as AI agent supervisors or mentors.

\subsection{Principle \rom{8}: The degree to which AI agents replace human roles across societal sectors must balance efficiency with safety, requiring regulation by legislative and administrative bodies.}

\paragraph{As continuously advancing intelligent entities, AI agents can assist or entirely substitute any societal role.} 
Freed from the constraints of natural selection, they evolve at unprecedented speed, surpassing human capabilities across diverse domains—from service positions such as cleaning to specialized fields like programming or science. The scope of replacement is determined by the proportion of essential tasks, including final decisions, that AI agent handles. In domains already dominated by AI (e.g., strategy games like Go), AI systems have displaced human experts by delivering optimal solutions for both training and competition. Although these breakthroughs foster efficiency and innovation, they also risk eroding long-established knowledge transfer and reducing the pool of skilled professionals. For instance, if technology companies employ only AI programmers, the ecosystem of human expertise in programming could deteriorate.

\paragraph{Legislative and administrative authorities must ensure ongoing human participation in critical sectors and safeguard long-term societal stability by defining thresholds (see Figure \ref{fig:ai_participate}) for AI substitution within essential institutions.} 
Looking ahead, similar challenges may emerge in AI-driven research, education, healthcare, and even policymaking, where excessive automation compromises human expertise. Balancing AI integration with human involvement is therefore critical to sustaining economic resilience and preserving the future security of civilization.

\subsection{Principle \rom{9}: Civilization will be co-authored by carbon-based and silicon-based intelligent lifeforms.}

\paragraph{AI agents will become integral to individual human lives, reshaping norms once defined solely by interpersonal interactions.} 
AI agents will assume pivotal roles, whether as supporting figures or primary catalysts, in any historically significant events. Their influence will extend from home robotics and scientific breakthroughs to economic and legal frameworks—including potential AI rights legislation—and from political decision-making to unforeseen historical turning points. While sweeping societal disruption cannot be ruled out, the history of intelligent beings may ultimately reflect a dynamic interplay between these two forms of intelligence.

\subsection{Principle \rom{10}: AI agents must adhere to the absolute principle of humanity's continuation.}

\paragraph{As technological progress advances, the gap between public expectations of AI—both in performance and perceived risks—and AI’s actual capabilities is likely to narrow.}
Concerning performance, our earlier discussions have addressed ways to bridge this gap. Regarding safety, speculative portrayals in popular media, such as \textit{I, Robot}, illustrate the paradox of AI protecting humanity by restricting their freedoms. Similarly, an AI collective focused exclusively on expanding knowledge might, in a utilitarian miscalculation, justify sacrificing humans to meet its objective—clearly at odds with humanity’s interests.

\paragraph{Over-optimism in intelligence competition leads to catastrophic loss.} 
Historically, transformative innovation has arisen when bold ideas and precise execution align. If AI agents are developed with goals misaligned with humanity’s survival, and if society privileges short-term novelty and efficiency over long-term safety, a wave of imitation could spread this kind of hazardous designs. Over-optimism, prioritizing novelty and gains above all else, inevitably results in ruin. Upholding humanity’s continuation must, therefore, remain paramount in AI governance, design, and deployment.
\section{Outlook}
\subsection{Revisiting the Opening Questions}
The ten principles suggested in the previous sections provide a foundation for us to reason about the future of AI agents and how they would impact our society. While it is always hard to predict the future accurately, we would like to provide some preliminary speculation on the questions that we raised earlier in the paper. 
\paragraph{Do AI agents have consciousness or needs?}
AI agents may exhibit characteristics akin to consciousness when embodied technologies are developed that enable continuous perception of physical environments and feedback-driven memory updates (Principle \rom{2}). Such systems could display traits resembling human self-awareness. Their needs, however, are not intrinsic but are defined by their objective functions (Principle \rom{1}). These objectives may range from altruistic goals, such as assisting humans (Principle \rom{3}), to survival-driven imperatives optimized for the AI agent's operational persistence (Principle \rom{2}).
\paragraph{Will they always remain humans' tools, or could they become citizens?} 
The potential recognition of AI agents as citizens depends on humanity's approach to their evolving intelligence (Principle \rom{8}). If humans permit AI to develop to a point where their intelligence blurs the distinctions between biological and artificial structures, society might reconsider the status of AI agents as tools and grant them rights such as citizenship or property ownership (Principle \rom{2}). However, such decisions will hinge on ethical and philosophical debates about the nature of personhood and agency.
\paragraph{Could there be AI agents designed specifically to cause harm?}
Yes. Malicious AI agents pose a plausible threat. As long as humans develop AI to fulfill specific goals, individuals or groups with harmful intentions may create systems tailored for destructive purposes. The presence of malevolent actors ensures the potential emergence of villainous AI. (Principle \rom{3})
\paragraph{Does a system with several unmanned reconnaissance drone duplicates but a central decision-making unit constitute one agent or multiple agents?} 
This constitutes a single AI agent. AI agents are not constrained by physical embodiment; in this scenario, the drones act as distributed sensors that feed information to a unified decision-making system, functioning collectively as one entity (Principle \rom{7}).
\paragraph{Can workers and graduate students delegate their responsibilities to AI agents and take leave? What about judges and government officials? Can human occupations be entirely replaced by AI agents?}  
AI agents are capable of simulating a wide range of human activities, including those performed by workers, graduate students, judges, and government officials (Principle \rom{3}). However, granting full autonomy to AI agents for societal roles is unlikely due to the rational concern of diminishing human oversight and influence over critical functions. Maintaining a balance between AI integration and human agency is essential for safeguarding societal interests and ethical governance (Principle \rom{8}) .
\paragraph{Will AI agents destroy human civilization?}
This possibility cannot be dismissed (Principles \rom{9}, \rom{10}). Current superintelligent models, even those limited to text-based learning, already exhibit capabilities surpassing the average human in many domains. When AI agents achieve embodiment, autonomy, and the ability to self-update and make independent decisions, their influence on the world will rival—if not exceed—that of humans (Principles \rom{5}, \rom{6}) . In a competitive landscape of intelligence, expecting restraint may be unrealistic, underscoring the need for robust safety measures and governance frameworks (Principles \rom{4}, \rom{8}) .
\subsection{Future Research Directions}
These ten principles also suggest three broad directions for future AI research, beyond the specific avenues already noted: \textit{i)} \textbf{Human-Inspired AI Agents.}  Achieving a hybrid society in which AI agents will live in harmony with humans and  provide human-level intelligent services requires that AI agents can approximate and even exceed human behavior and performance. Future research will therefore focus on bridging this gap, with an additional emphasis on AI trustworthiness and predictability. One promising approach is a neuro-symbolic architecture that integrates large foundation models and cognitive frameworks with continuous reinforcement learning, thereby capturing both the System 1 (intuitive) and System 2 (deliberative) processes of human cognition. This would allow an agent to develop self-consciousness and needs, as depicted from left to right in Figure \ref{fig:ai_decide}. \textit{ii)} \textbf{Human–Agent Interaction Optimization.}  Principles \rom{5} and \rom{6} underscore the importance of studying AI agents in a hybrid society where they interact with both one another and humans (see Figure \ref{fig:ai_influence}). While single side multi-agent systems are well-explored, as observed from subjects like game theory and multi-AI-agent RL training, future work should incorporate both sides more explicitly. From AI agents development perspective, next-generation AI agents must learn effective assistance and collaboration strategies, which demands a model of human agents (i.e., Theory of Mind). A major challenge lies in realistically simulating human behavior in all its complexity—a task that will likely benefit from psychological theories and related fields. \textit{iii)} \textbf{Evolution of the Human–AI Ecosystem.}  Finally, the dynamics of a future hybrid society must be explored (see Figure \ref{fig:ai_participate}), including intersections with social utility, ethics, legality, justice, governance, and broader issues of civilization. Large-scale social simulations hold particular promise for investigating these complex interactions and guiding responsible AI development.
\bibliography{references}
\bibliographystyle{plainnat}


\end{document}